\documentclass{article} 
\usepackage{iclr2023_conference,times}
\usepackage{graphicx}
\usepackage{booktabs}
\usepackage{multirow}
\usepackage{caption}
\usepackage{inputenc}
\usepackage{caption}
\usepackage{subcaption}
\usepackage{wrapfig}
\usepackage{amsmath,enumerate}
\usepackage{siunitx}

\usepackage{amsmath,amsfonts,bm}

\newcommand{\EQCOMMA}{\ensuremath{\,\textrm{,}}}
\newcommand{\EQDOT}{\ensuremath{\,\textrm{.}}}

\def\eqref#1{equation~\ref{#1}}

\def\1{\bm{1}}

\DeclareMathAlphabet{\mathsfit}{\encodingdefault}{\sfdefault}{m}{sl}
\SetMathAlphabet{\mathsfit}{bold}{\encodingdefault}{\sfdefault}{bx}{n}

\usepackage{hyperref}
\usepackage{url}
\hypersetup{
  colorlinks,
  allcolors=.,
}

\title{Interpretable Reinforcement Learning via Neural Additive Models for Inventory Management}

\author{Julien Siems\thanks{Corresponding author}, Maximilian Schambach, Sebastian Schulze, Johannes S.\ Otterbach\\
Merantix Momentum GmbH\\
Max-Urich-Str.\ 3\\
13355 Berlin, Germany \\
{\small \texttt{\{julien.siems,maximilian.schambach,sebastian.schulze,}} \\
{\small \texttt{johannes.otterbach\}@merantix.com}} \\
}

\iclrfinalcopy %
\begin{document}
\maketitle
\begin{abstract}
The COVID-19 pandemic has highlighted the importance of supply chains and the role of digital management to react to dynamic changes in the environment. In this work, we focus on developing \emph{dynamic} inventory ordering policies for a multi-echelon, i.e. multi-stage, supply chain. Traditional inventory optimization methods aim to determine a \emph{static} reordering policy. Thus, these policies are not able to adjust to dynamic changes such as those observed during the COVID-19 crisis. On the other hand, conventional strategies offer the advantage of being interpretable, which is a crucial feature for supply chain managers in order to communicate decisions to their stakeholders.
To address this limitation, we propose an interpretable reinforcement learning approach that aims to be as interpretable as the traditional static policies while being as flexible and environment-agnostic as other deep learning-based reinforcement learning solutions. We propose to use Neural Additive Models as an interpretable dynamic policy of a reinforcement learning agent, showing that this approach is competitive with a standard full connected policy. Finally, we use the interpretability property to gain insights into a complex ordering strategy for a simple, linear three-echelon inventory supply chain.
\end{abstract}

\section{Introduction}\label{sec:introduction}
The impact of the COVID-19 pandemic has brought the importance of reliable supply chains to everyone's attention. Empty supermarket aisles and medication shortages have shown the general public the fragility of modern supply chains. From a consumer's perspective, it is vital that supply chains are robust enough to handle unexpected changes in demand, supplier issues, and transport disruptions, making it essential for supply chain optimization to address these concerns. 
Supply chains are a central topic within operations research and are typically broken down into Supply Chain Design (SCD), Supply Chain Planning (SCP), and Supply Chain Execution (SCE)~\citep{wenzel2019literature}. 
SCD describes the process of planning the locations, capacity and general demand policy of a supply chain, while SCP describes the medium-term strategy, such as production cycles and logistics. 
Finally, SCE deals with the supply chain management and its control. From this it is clear that the robustness of a supply chain may be influenced at different moments in its life cycle. The focus of this work lies on inventory optimization, a subpart of SCE, using dynamic yet interpretable policies. 

Since early seminal works by~\citet{harris1913many}, inventory optimization has been a focus of operations research. It is now widely used in enterprise resource-planning systems that help operate most manufacturing and distribution companies. Traditionally, inventory optimization methods aim to construct a \emph{static} policy, based on minimum inventory levels and reorder quantities. On the one hand, these static policies can easily be implemented in a fixed process and, on the other hand, are easy to interpret and communicate to supply chain stakeholders. However, these methods rely on assumptions such as linearity of costs or expected demand distribution~\citep{snyder2019fundamentals, mula2010mathematical} and cannot easily be adjusted to changes in a dynamic environment -- not to mention drastic short-term disruptions such as those caused by the \textsc{Covid}-19 pandemic. As a consequence, static policies are often suboptimal with respect to cost and efficiency.

With the advent of modern computational methods, the focus has since shifted to construct more complex policies. These can be based on simulation tools, like the static {OTD-Net}~\citep{odtnet} simulator, Monte-Carlo-based discrete event simulators such as {SimPy}~\citep{simpy}, or optimal control techniques. Within the latter category, Reinforcement Learning (RL) can be used to find \emph{dynamic} policies that do not require assumptions about the environment. Paired with a broad range of simulatable dynamic supply chain scenarios, these policies can be made resilient against disruptions of the supply chains, or can be used in conjunction with continual learning to dynamically react to new set points of the supply chains. However, state-of-the-art RL methods rely on Neural Networks (NNs) to represent policies making their interpretation difficult. 

To overcome these limitations, we propose an \emph{interpretable} RL approach based on Neural Additive Models (NAMs)~\citep{agarwal2021neural}. Using a NAM to represent the agent's policy, we can leverage its interpretability to gain insights into the learned features of the policy, closing the gap towards static and interpretable approaches. At the same time, their functional form is highly flexible to be on par with other deep learning-based RL approaches when learning environment-agnostic dynamic policies. 

To the best of our knowledge, this paper provides the first application of full end-to-end \textit{interpretable} neural policies to solve challenging, dynamic optimization problems in an Inventory Optimization benchmark. Thus, our approach explicitly contributes to one of the grand challenges of interpretable machine learning set out by~\citet{rudin2022interpretable}.

\section{Related Work}\label{sec:related_work}

\paragraph{Inventory Optimization}
Traditionally, inventory optimization has been approached by finding a static base-stock policy~\citep{snyder2019fundamentals}. This strategy reorders whenever the inventory stock levels fall below a certain threshold.
The base-stock levels are commonly determined via Mixed-Integer Linear Programs (MILP). Due to its simplicity, this approach offers the advantage of being interpretable and the resulting policy can easily be communicated to relevant stakeholders. We refer to~\citet{snyder2019fundamentals} for a thorough introduction to static-policy inventory optimization.

RL has been proposed to make inventory optimization more responsive towards changes in the environment, due to its flexible dynamic policies. One of the earliest works on RL in inventory optimization is by~\citet{giannoccaro2002inventory} which models replenishment orders via tabular Q-learning to optimize the profit across the supply chain. Subsequently,~\citet{mortazavi2015designing} use Q-learning via NNs to maximize the profit of a four-echelon supply chain.
These early works address single- rather than multi-product supply chains and consider only a single retailer, while usually warehouses supply their products to various retailers. \citet{sultana2020reinforcement} deal with these challenges by optimizing supply chains of 50 to 1000 products with three retailers.

\citet{hubbs2020or} proposed the first RL benchmark for operations research including environments for inventory optimization. It is based on the OpenAI Gym environments~\citep{brockman2016openai} making it easy to benchmark a variety of RL algorithms via RLlib~\citep{liang2018rllib} or Stable Baselines~\citep{stable-baselines, Raffin2021StableBaselines3RR}.
The only other work we are aware of investigating a similar interpretable inventory optimization is~\citet{bravo2020mining}. The authors propose to identify optimal order strategies through collecting data generated by an optimal policy found via Dynamic Programming. It then frames the decision whether to reorder as a classification problem and fits a separate model to approximate the reorder quantity in the case a reorder takes place.

For a comprehensive introduction to RL-based inventory optimization, we refer the interested reader to~\citet{boute2021deep}.

\paragraph{Interpretable Reinforcement Learning}
To distinguish between \textit{interpretability} and \textit{explainability}, we follow the definitions by~\citet{glanois2021survey}. 
They define \textit{interpretability} as an architectural design of a machine learning model that allows humans to intuitively understand the model's decision making.
In contrast, \textit{explainability} is a post-hoc process that seeks to justify the prediction of a model after the fact.
Following this definition, explainable machine learning can be viewed as a collection of model-agnostic methods to reason about the model's decision whereas interpretable machine learning utilizes specific model architectures to enable human inspection of its decisions.

While interpretable and explainable artificial intelligence method have been gaining attention in recent years, research has mostly focused on applications such as computer vision or natural language processing using supervised or self-supervised architectures.
In contrast, interpretability and explainability in RL have attracted only little attention. In this context, \citet{verma2018programmatically} represent policies using a high-level description language which is found via a custom search method. As a result of the description language, the derived policies may be verified using traditional program verification tools. Their policy provides more stable steering directions in a car steering task and is much more robust to noise as compared to previous approaches.
\citet{coppens2019distilling} develop a method to learn soft-decision trees via distillation of a NN agent. Using the Mario benchmark environment~\citep{karakovskiy2012mario}, they record the state-to-action probability mapping of a trained NN policy. Subsequently, they fit a soft-decision tree on the collected data. We refer to~\citet{glanois2021survey} for an in-depth review of interpretable RL.

\section{NAM-PPO for Learning Interpretable Supply Chain Policies}\label{sec:method}

\begin{figure}
    \centering
    \includegraphics[width=0.8\textwidth, height=190pt]{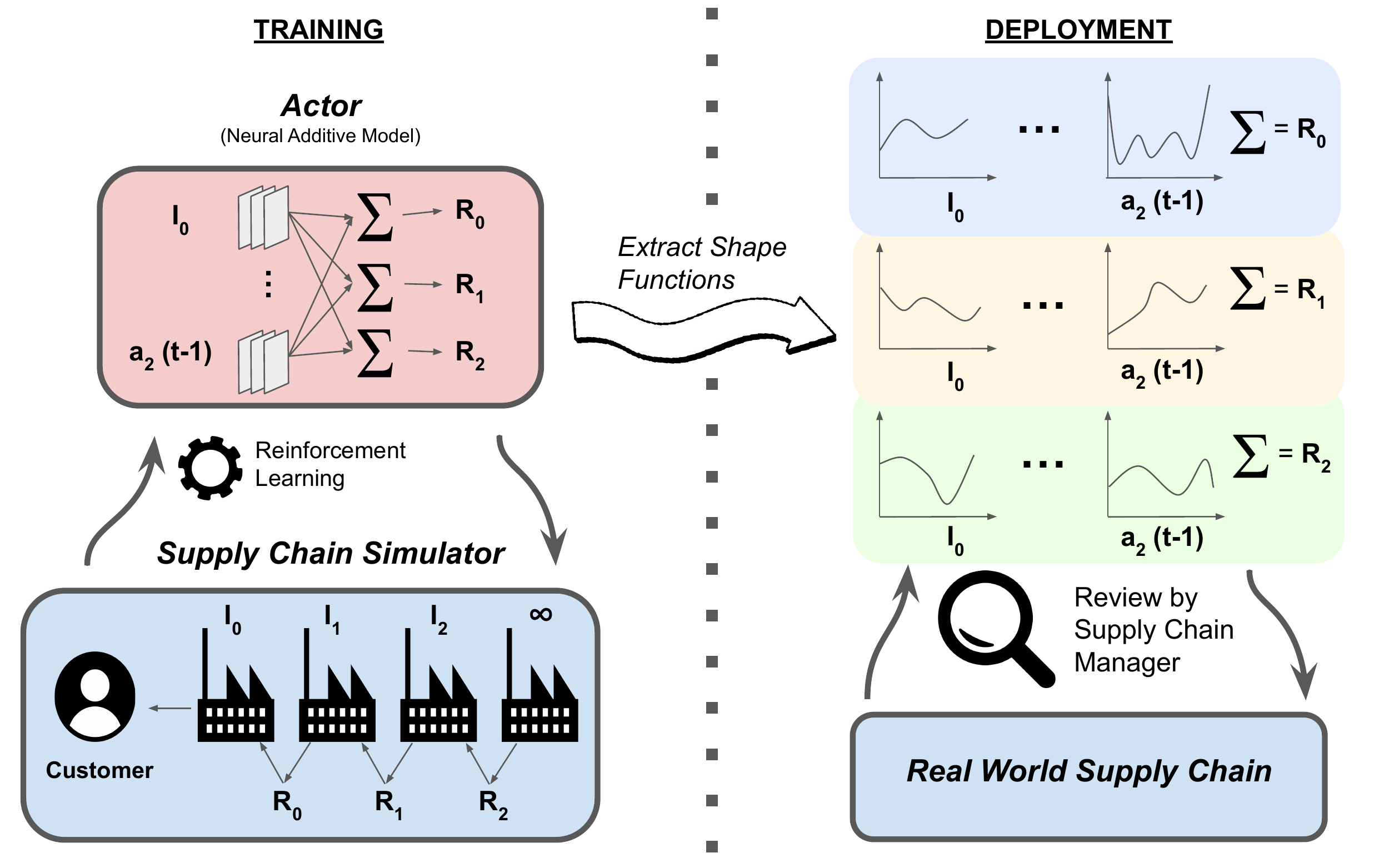}
    \caption{(Left) An agent, modeled as a NAM, is optimized via RL to produce the reorder amount $R_i$ for each stage $i$ of a multi-echelon supply chain in a simulation. The agent observes the current state of the inventory $I_i$ at each stage $i$ and recalls the last actions to determine the next one. (Right) Once  deployed, the shape functions may be extracted for each output of the NAM, since each feature contributes independently to the target. The shape functions allow the Supply Chain Manager to inspect how each feature in the observation space impacts the decision of the reorder amount.}
    \label{fig:method:overview}
\end{figure}

At a high level, our setup consists of three key components: the environment simulator providing us with a corresponding supply chain, an RL training framework to determine a dynamic supply chain policy, and an interpretable model to parameterize the policy. We visualize these components in Figure~\ref{fig:method:overview}.

\paragraph{Supply chain environment.} To keep the environment simple and focus on the use of interpretability when learning neural policies, we restrict the problem to linear-chain multi-echelon inventory optimization (MEIO) as depicted in Figure~\ref{fig:method:overview}.

To solve the ordering policy of this and similar supply chains, the supply chain manager typically implements a base-stock policy: Each facility keeps a minimum stock of the items and places a restocking order with their supplier whenever the inventory drops below the base-stock threshold. While this policy is able to meet the requirements of being able to fulfill the orders at the retailer level, it is apparent that a static base stock policy can also be suboptimal, as, e.g., a large base-stock in times of low demand leads to a high holding cost overhead. A dynamic and at the same time simple-to-understand policy could hence help in reducing the supply chain costs.

\paragraph{Training a dynamical supply chain using Proximal Policy Optimization.} Since its inception, Proximal Policy Optimization (PPO)~\citep{schulman2017proximal} has quickly become the workhorse for many RL applications. It has been successfully scaled to large games such as playing Dota 2~\citep{berner2019dota} and Multi-Agent Hide \& Seek~\citep{Baker2019EmergentTU}, but also applications such as automated stock trading~\citep{Liu2020FinRLAD} or Neural Architecture Search~\citep{Zoph2017LearningTA}. PPO is an on-policy algorithm that improves Trust Region Policy Optimization (TRPO)~\citep{schulman2015trust} by removing the computational bottlenecks of the latter. Specifically, we use the clip version of PPO which regularizes the update size of the policy's parameters based on the current policy and the advantage given the current state by clipping the reward signal,
\begin{align}
    \mathcal{L}^\textrm{PPO}(\theta, \theta_\textrm{old}) = \mathbb{E}\left[\min\left(\frac{\pi_\theta(a_t|s_t)}{\pi_{\theta_\textrm{old}}(a_t|s_t)}A^\pi(s_t), \textrm{clip}\left(\frac{\pi_\theta(a_t|s_t)}{\pi_{\theta_\textrm{old}}(a_t|s_t)}, 1-\varepsilon, 1+\varepsilon \right)A^\pi(s_t)\right) \right] \EQDOT
\end{align}

Our implementation of PPO is provided by {Stable-Baselines3}~\citep{Raffin2021StableBaselines3RR}, which uses distinct networks for the policy $\pi_\theta(a|s)$ as well as the value function $v_\phi(s)$. The advantage values are generated using Generalized Advantage Estimation~\citep{Schulman2015HighDimensionalCC}, $A^\pi(s) = R - v_\phi(s)$, where $R$ is the discounted reward. We note that the advantage function's sole purpose is to support the computation of the policy gradient during training. In particular, it has no bearing on the agent's action selection which is fully governed by the policy network. 

Within the standard RL approaches, the policy is parameterized using fully connected multi-layer perceptrons (MLP). As a consequence, interpreting and explaining the action selection of the agent is challenging, thus limiting their utility in real world scenarios that require stakeholder management, such as in critical supply chains. To address this issue, we make use of interpretable models to replace the opaque MLP policy networks with interpretable Neural Additive Models (NAM)~\citep{agarwal2021neural}. In section~\ref{sec:experiments} we compare the standard MLP-PPO to its interpretable NAM-PPO counterpart. Here we make use of the fact that {Stable-Baselines3} allows us to separate policy and value networks, as we use the same value network design for both policy setups for fair comparison.

\paragraph{Neural Additive Models as interpretable supply chain policies.} \label{sec:method:nam}As pointed out before, our goal is to design an interpretable policy network to aid communicating the agent's decision to supply chain managers. We use NAMs, a form of Generalized Additive Model (GAM)~\citep{hastie1987generalized,hastie2017generalized}, whose functional form is given by
\begin{equation}
    \hat{y} = \beta + \sum\nolimits_{i=1}^N f_i(x_i) \EQCOMMA
\end{equation}
where $(x_i)_i \in \mathbb{R}^N$ is a feature vector of $N$ scalar-valued entries, $f_i: \mathbb{R} \to \mathbb{R}$ are univariate, continuous functions describing the individual feature contribution, $\beta \in \mathbb{R}$ is a global offset or bias term, and $y$ denotes the target variable.
GAMs are an extension of linear regression and add non-linear feature contributions while retaining the ease of interpretability. 
The model is interpretable by tracing its so-called \emph{shape functions}, i.e.\ the graphs $\left\{\big(x_i, f_i(x_i)\big), x_i \in \mathcal{D}_i\right\}$, in the respective domain of interest $\mathcal{D}_i \subset \mathbb{R}$.
The shape functions describe the contribution of the corresponding features \emph{exactly} as the prediction is given by a simple point-wise addition of feature contributions.
This way, domain experts or other stakeholders can inherently understand the model's decision making or gain insight into the data the model was trained on to validate or expand expert knowledge.

Until recently, the feature contributions $f_i$ were modelled by splines and then replaced by boosted decision trees which perform more favorably in many applications~\citep{lou2012intelligible,lou2013accurate,caruana2015intelligible}.
NAMs follow the same idea but express the single-feature contributions $f_i$ using MLPs. 
Hence, NAMs are differentiable by construction and benefit from recent advances in deep learning and hardware acceleration. Furthermore, the shape functions can also be used to deploy the model in an efficient way, e.g., via look-up tables. In that case, the actual NN implementation of the NAM can be traced and discarded after training.
Note that the input features and target are standardized for numerical stability, for example by using a min-max- or z-scaler.

Finally, NAMs are easily adaptable to multi-task learning with weight sharing between tasks: Each task target $y_t$, $t=1, ..., T$, is modelled as
\begin{equation}
    \hat{y}_t = \beta_t + \sum\nolimits_{i, s} w_{t, i, s}f_{i, s}(x_i) \EQCOMMA
\end{equation}
where $s=1,...,S$ denotes a subnet index, $f_{i,s}$ are parametrized as MLPs, and the scalar weights $w_{t,i,s}$ are also trainable. Note, that each task-specific feature contribution or shape function $f_{t,i}$ is effectively given by a weighted sum
$f_{t,i}(x_i) = \sum\nolimits_{s} w_{t, i, s}f_{i, s}(x_i)$
of subnet contributions $f_{i,s}$. Using weighted subnets as opposed to training independent feature contributions for each task allows for efficient parameter sharing of feature contributions across tasks. For more technical details, we refer to the original work on NAMs~\citep{agarwal2021neural} and the follow-up work~\citep{chang2021node}.

\section{Experiments \& Discussion}\label{sec:experiments}

In the following, we provide details on our hyperparameter optimization and evaluation strategy in Section~\ref{sec:exp-setup}. 
In Section~\ref{sec:exp:benchmark_comparison}, we compare two PPO-based setups using a standard MLP policy and an interpretable NAM policy head-to-head and leverage the interpretability to provide insights on the learned strategies. 
We refer to these setups as MLP-PPO and NAM-PPO, respectively.

\subsection{Experimental Setup}\label{sec:exp-setup}

\paragraph{MEIO in OR-Gym.} Our environment is provided by OR-Gym~\citep{hubbs2020or}. More specifically we use \textsc{InvManagement-v0}, which provides an MEIO problem consisting of a linear chain with three echelons. Each echelon is characterized by its storage capacity, holding and fulfillment price, and demand fulfillment lead time. In addition to the previous parameters, the environment includes an order backlog, i.e. if an echelon is unable to fulfill an order at a given time, it is added to a backlog and fulfilled at a later time at reduced price instead of writing off the unfulfilled orders. The randomness in the environment is introduced via a stochastic customer demand at the lowest echelon, i.e. the retailer, of the supply chain. The retailer tries to fulfill the customer demand from its inventory and is able to post a replenishment order to its supplier which takes a certain lead time to arrive. Similarly, every supplier has an inventory and can order from a single upstream supplier. The first supplier in the supply chain has no source to order from, but accesses an infinitely large inventory.

During the experiments we found two key settings in the environment to enable the agent to learn meaningful policies. First, we note that we deal with finite time horizons and hence can remove the reward discounting factor. This change is critical for the agent to recognize the long-term benefit of replenishing stocks, as otherwise the initial (undiscounted) order and inventory prices outweigh the realizable downstream revenue and learned policies degenerate. Second, we add additional stochasticity by randomizing the initial inventory using samples from a Gaussian distribution centered at the default initial OR-Gym inventory levels (I0: 100, I1: 100, I2: 200) and a standard deviation of 50 units.

As inputs to the RL agent, we use the OR-Gym default observation space containing the current inventory state $I_i$ in each echelon (3 values) and the actions of the last 10 time-steps (30 values). Hence, the observation space is 33-dimensional in total. The action space at each time-step is given by the agent's reorder quantity per echelon, resulting in a 3-dimensional vector. Finally, the reward is defined as the revenue generated by fulfilling customer demand through the retailer, minus the inventory cost at each echelon. For more details we refer to the OR-Gym work by~\citet{hubbs2020or}.

\paragraph{Hyperparameter Configuration and Optimization.} There are multiple hyperparameters used throughout the experiments.
As the evaluation is restricted to a single example supply chain, the number of target tasks (i.e., the dimension of the action space) for the NAM and MLP policies is fixed and equal to 3.
For both considered policy network architectures, the PPO critic is a simple MLP with two layers of fixed width.
Furthermore, we utilize ELU activation functions throughout the policy and critic networks as we have found the one originally proposed in the NAM context, the so-called Exp-centered Unit (ExU)~\citep{agarwal2021neural}, as well as conventional Rectified Linear Units (ReLU) to be unstable. We leave the investigation of normalization layers to mitigate instabilities to future work. The depth and width of the MLPs, for MLP-PPO as well as NAM-PPO, are determined through hyperparameter optimization.
Note, that both the actor policy as well as the critic operate directly on the raw features, e.g.\ no shared feature extractor is used.
While this might be less parameter-efficient, it is required to achieve interpretable feature contributions using the NAM actor.

We optimize the hyperparameters of the used optimizer, PPO, and the policy networks using Random Search~\citep{bergstra2012random}. 
In total, 30 hyperparameter configurations are sampled. We evaluate each configuration by training three policies based on different random initialisations and averaging their cumulative reward over 60 steps in a validation environment with a fourth, fixed seed. The used hyperparameters and corresponding search spaces for optimization are presented in detail in Appendix~\ref{app:sec:hyperparameter_optimization}. All remaining hyperparameters of PPO follow the {Stable-Baselines3} defaults.

\paragraph{Training and Evaluation.}\label{sec:exp:evaluation}
We follow the recommendations of~\citet{agarwal2021deep} for a reliable evaluation of RL policies and use the interquartile mean (IQM) with 5\% and 95\% bootstrap confidence bounds for tabular comparisons of different methods across seeds and agent rollouts.  More specifically, we retrain the incumbent of the hyperparameter optimization using 20 different random seeds and roll out each of the resulting agents 50 times. The IQM is hence computed based on 1000 evaluations in total. Each agent is trained and evaluated for an episode length of 60.

\begin{figure}[t]
     \hfill
     \centering    
     \hfill
       \begin{subfigure}[b]{0.3\textwidth}
         \centering
         \includegraphics[width=\textwidth]{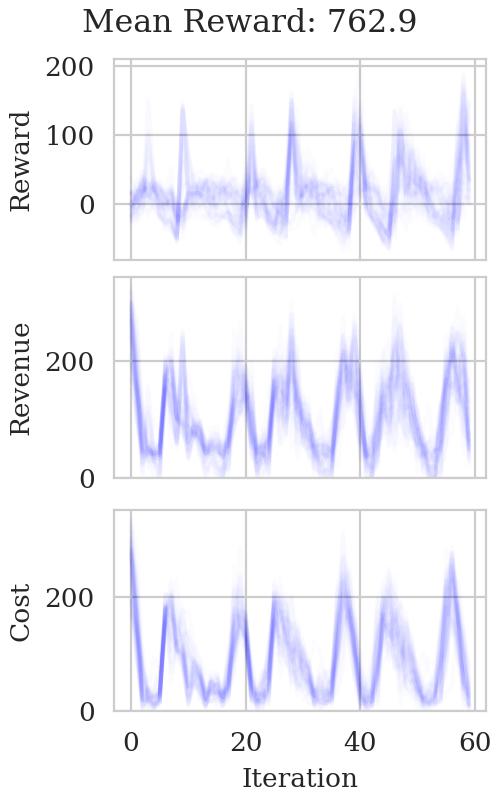}
         \caption{}
         \label{fig:exp:rollout_per_step}
     \end{subfigure}
     \hfill
     \begin{subfigure}[b]{0.475\textwidth}
         \centering
         \includegraphics[width=\textwidth]{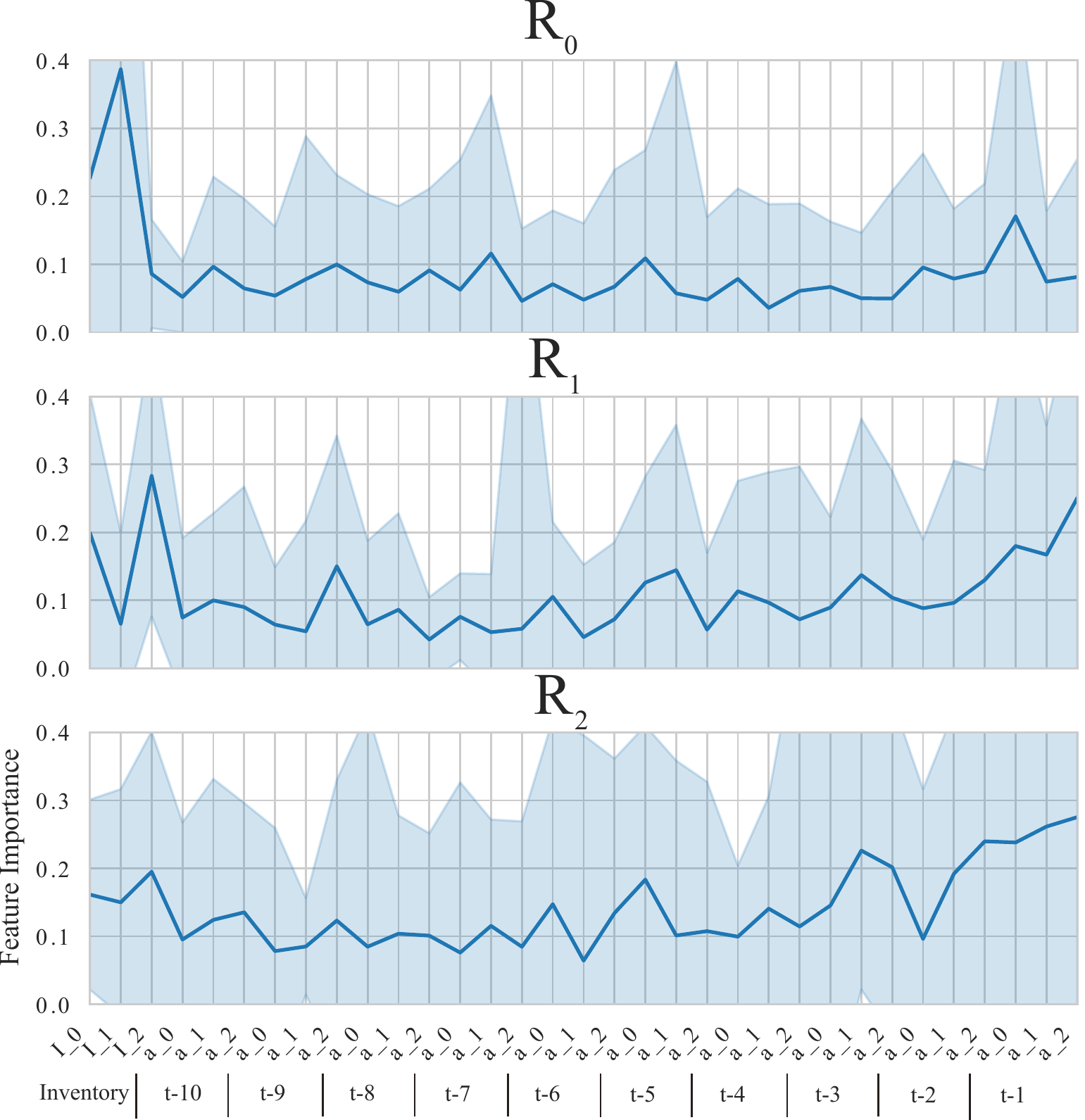}
         \caption{}
         \label{fig:exp:nam_feature_importance}
     \end{subfigure}
     \hfill
     \hspace{5mm}
    \caption{(a) Example per-step reward trajectory of the NAM agent. Shown are the 50 rollouts for one of the 20 trained agents. (b) Feature importance in the observation space, the median importance per feature across the 20 model trainings is shown, shaded is the standard deviation.}
\end{figure}

\subsection{Results}\label{sec:exp:benchmark_comparison}

For the default training and evaluation setup, we find that the IQM of the NAM-PPO is 792.69, with a confidence interval of [779.96, 804.90], and 730.27  for MLP-PPO with a confidence interval of [709.12, 750.38]. 
Hence, NAM-PPO performs slightly better than the standard MLP-PPO in the considered example.
This is remarkable as the NAM agent does not consider higher-order feature interaction and can, by design, model less complex policies as compared to an MLP agent.
In particular, a NAM-based policy can only linearly combine the different inventory state contributions or those of the previous reordering quantities which can, in principle, interact in highly complex ways in the MLP-based policy.
Hence, these initial results encourage further investigation into NAM-based RL.

Example rollouts of the NAM-PPO policy for one of the 20 seeds is shown in Figure~\ref{fig:exp:rollout_per_step}. We observe that the NAM policy learns to produce an oscillating reward/profit sequence. 
Ostensibly, the agent waits for inventories to empty (decreasing cost) before replenishing stocks (jump in cost) imitating a classic base-stock policy, which -- given the simplicity of the considered supply chain -- may well be optimal. However, it is difficult to say whether the learned policy is based on inventory thresholds, a schedule, or other heuristics.

With the help of NAM-based policy networks, we can assess the relevance of different environment aspects for overall decision making, by averaging and comparing the respective absolute feature contributions across a set of states of interest. We can further investigate the relationship between agent behavior and any individual environment feature by examining the matching shape function relating variations in the feature value to increased or decreased task (action) contribution.

Figure~\ref{fig:exp:nam_feature_importance} depicts the feature importance of each variable in the observation space estimated by the NAM-based policy for the individual actions $R_0$, $R_1$, and $R_2$, denoting the reordering quantities of the echelon inventories $I_0$, $I_1$, and $I_2$, respectively.
The actions from previous time steps are denoted as $a_i(t-c)$, $i=1,2,3$, where $c$ denotes the time delay. For example, $a_2(t-1)$ denotes the reordering quantity of echelon $2$ in the previous time step.
For $R_0$ (top row), the reordering amount at the producer closest to the customer, we see that inventories $I_0$ and $I_1$ are the most important features. This makes sense as $I_1$ determines how much can be reordered by the producer with inventory $I_0$ from the preceding producer in the supply chain. 
Similarly, $R_1$ (middle row) is influenced most strongly by the current inventory $I_2$. Interestingly, the current level $I_1$ of the intermediate supplier himself is of little consequence compared to the retailer stock $I_0$ which is an indicator for future orders.
For $R_2$ (bottom row), the reordering quantity from the producer with infinite inventory, the current inventory states $I_i$ are almost negligible whereas the actions taken in the previous time-steps are most important.
This seems reasonable, as it likely is important to keep track of the previous actions taken for $R_2$, which may indicate what quantity of goods is still in the pipeline and will arrive at a later time, due to lead times in the delivery. 
This trend seems to be observable for all reordering quantities, in particular for those with increasing distance to the customer. However, more detailed investigations using more complex supply chains are required.

\begin{figure}[t]
    \centering
    \includegraphics[width=0.8\textwidth]{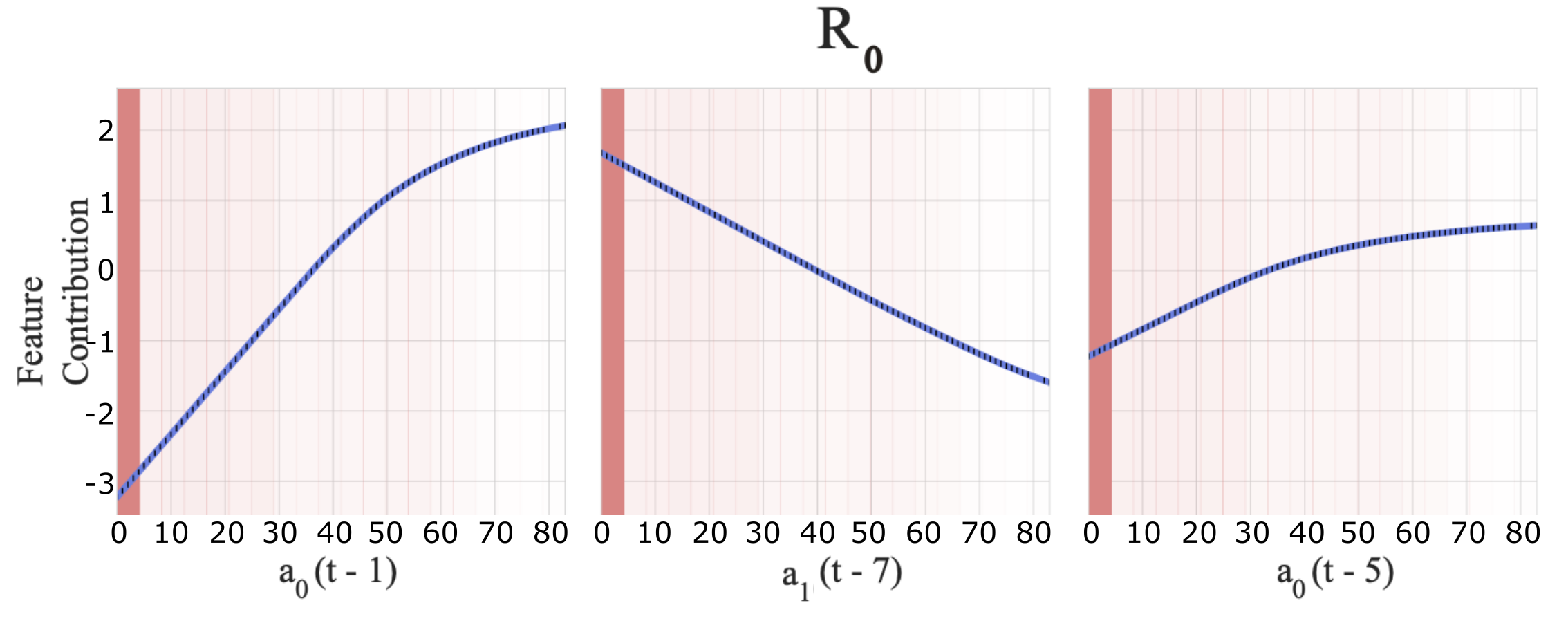}
    \caption{Example shape functions for target action $R_0$ corresponding to the reorder quantity of the first stage in the supply chain extracted from one of the trained NAMs in Section~\ref{sec:exp-setup}. Shown are the three features with highest feature importance in descending order from left to right. The shaded areas denote the data density in this interval, which, in the RL case, corresponds to how often the agent saw the corresponding feature in this state.}
    \label{fig:shape_functions}
\end{figure}

Finally, the interpretation using feature importance, while offering first insights into the policy's behavior, is limited in its expressiveness.
To this end, the shape functions of the NAM-based policies can be visualized as shown in Figure~\ref{fig:shape_functions}.
Here, we show the top-3 reordering features, i.e. it excludes the inventory observations, of the reordering history $R_0$ of the learned NAM-based policy. Recall, the shape functions exactly describe the contribution of the corresponding feature to the target, here, the reordering quantity $R_0$.
The background shading indicates data density for certain values and can be seen as a proxy for the state-visit distribution of the deployed policy. This offers first insights into the scenarios encountered by the agent and may also serve as an indicator for relevant parts of the state space.
In our example we see that order levels $R_0$ typically remain low (order quantity $0$--$15$), with sporadic increases (up to $60$). The main driver for these increases are immediately preceding orders $a_0(t-1)$ which have an amplifying effect, which we hypothesize hints to the fact that high preceding orders may indicate a backlog of unfulfilled demand. This effect saturates to avoid excessively large single orders. To avoid continuously overstocking, this contribution is counteracted by historic order levels further back $a_1(t-7)$ where larger orders have a dampening effect on orders at echelon $1$ to compensate for overstocking for a reduced demand. This type of analysis provides a simple, yet effective tool into understanding the complex behavior of neural policies in RL, that is also accessible to non-practitioners.

We provide additional preliminary evaluations in Appendix~\ref{app:further-evaluation}.
In particular, we analyze the temporal stability and extrapolation of the derived policies by evaluating them for more environment steps than they encountered during training. Moreover, we evaluate the robustness of both policy variants towards disruptions in the demand distribution.

\section{Conclusion}\label{sec:conclusion}

We proposed an interpretable reinforcement learning approach to multi-echelon inventory optimization. Our approach combines the interpretability of static inventory optimization methods with the flexibility of neural policies obtained trained via reinforcement learning with proximal policy optimization (PPO). We have demonstrated that Neural Additive Models (NAM) are good candidates to induce interpretability to an agent's policy. In a benchmark environment, the resulting policy produces results on-par with a neural black-box architecture such as a simple feed-forward network. We use the resulting feature importances and shape functions of the NAM policy to investigate the behavior of the agent in an example rollout and demonstrate how the complex interaction of ordering in a multi-echelon environment can be broken apart. This clearly demonstrates the use of interpretable policies to communicate dynamic policies to relevant stakeholders. 

This work opens up the study of many real-world applications of interpretable reinforcement learning for pressing supply chain and other operations research problems in the current economy. In order to unlock the full potential of NAM-PPO for these problems, we plan to analyze the addition of higher-order NAM features to the policy as well as extending the action space for more fine-grained supply chain control in future works. Goals are to understand long-time generalization, disruption robustness, as well as managing more complex supply chains. 

\section*{Acknowledgements}
This work has received funding from the German Federal Ministry for Economic Affairs and Climate Action as part of the ResKriVer project under grant no. 01MK21006H. We thank Ziyad Sheebaelhamd for in-depth discussions on Reinforcement Learning related problems. We thank Johanna Kim Kippenberger, John-Christopher Maleki and Michael Dominik Görtz for their fruitful discussions and insights on supply chain optimization.

\bibliography{main}
\bibliographystyle{iclr2023_conference}

\appendix
\clearpage
\section{Appendix}

\subsection{Further Evaluations}\label{app:further-evaluation}

\paragraph{Temporal stability}
    
\begin{figure}[!h]
\centering
\includegraphics[width=0.6\linewidth]{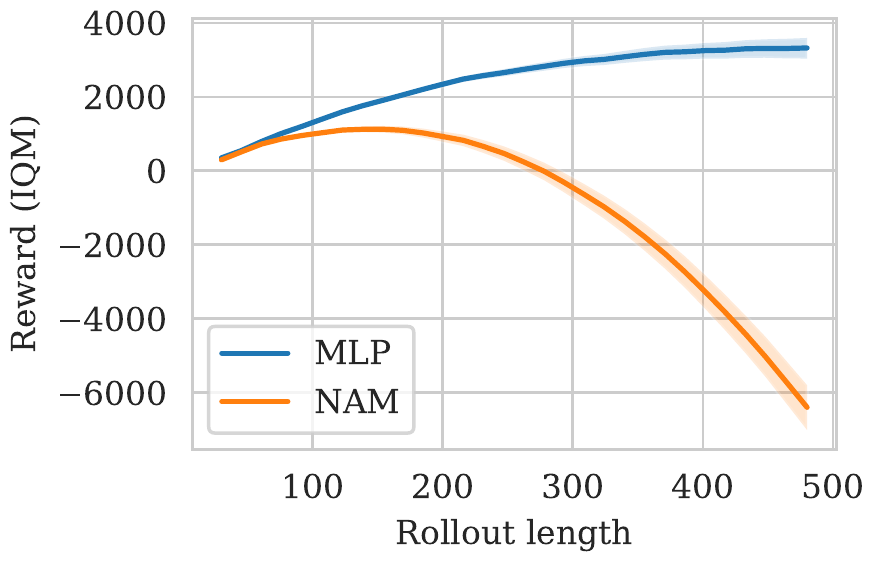}
\caption{Robustness of the learned MLP- and NAM-PPO policies towards increasing the roll-out length in the environment.}
\label{fig:rollout_length_analysis}
\end{figure}

We compare the temporal stability of a NAM-based policy with that of an MLP-based policy. 
Recall, the policies investigated in Section~\ref{sec:exp:benchmark_comparison} were trained for an episode length of 60 environment steps. Here, we vary the episode length between 30 and 420 environment steps and analyze whether the same policies continue to be profitable, i.e., if the cumulative reward remains positive. The results are shown in Figure~\ref{fig:rollout_length_analysis}. 
    
We find that the NAM policies generalize worse to an increased episode length than standard MLP policies. 
In fact, the NAM policy fail to remain profitable in the long run while the MLP policy remains stable and profitable.
We hypothesize that this may be due to the NAM's neglect of higher-order feature interactions which could be necessary to produce a stable policy.
However, this remains speculative and further works are necessary to investigate the long-term behavior of trained NAM-PPO policies.
    
\paragraph{Robustness towards disruptions in customer demand}

Next, we evaluate the robustness of the trained MLP- and NAM-PPO agents under sudden changes in customer demand, similarly to how they occur in a crisis. 
At time step $c$ in the environment, we add a disruption customer demand distribution to the default customer demand distributions.
The default distribution is modelled as a Poisson distribution with parameter $\lambda_{\textrm{s}}=20$.
The disruption distribution is also modeled as a Poisson distribution with $\lambda_{\textrm{d}}=s_{\textrm{d}} \cdot \lambda_{\textrm{s}}$, where $s_{\textrm{d}}$ denotes the disruption strength.
The samples from the crisis distribution are exponentially attenuated to zero over time with an attenuation factor of 0.8. Therefore, our disrupted demand changes suddenly and subsequently returns to the original demand distribution.

In all our experiments we set the time step $c$, at which the demand distribution changes, to be half the total number of environment steps in the experiments. This gives the policy enough time to stabilize (hence not purely relying on the initial inventory) as well as time to potentially recover.

\begin{figure}
\centering
\includegraphics[width=0.8\linewidth]{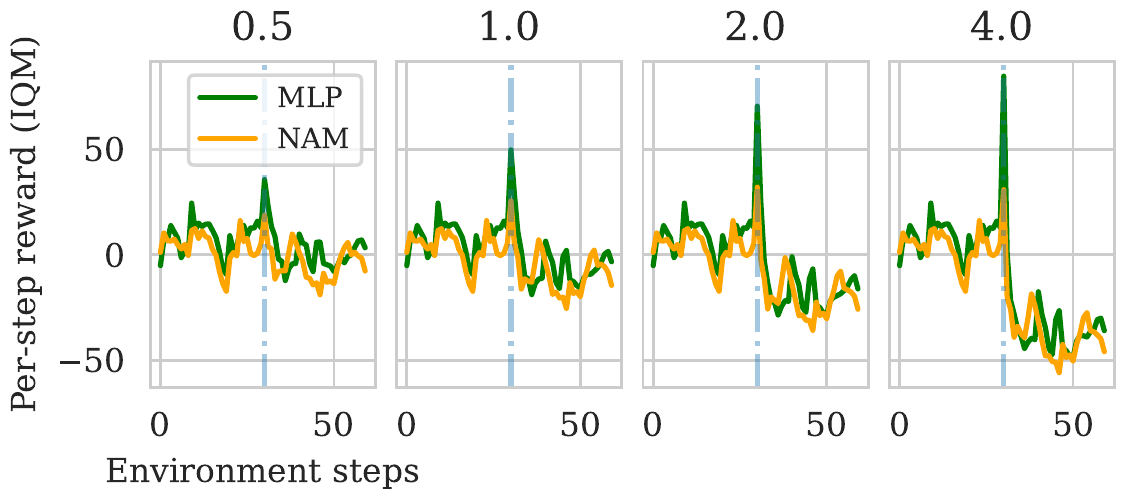}
\caption{IQM of the per-step reward over an environment rollout for different disurption strengths. The dashed vertical line indicates the time step of the disruptive demand distribution change. The disruption factor $s_d$ is denoted above each subplot.}
\label{fig:customer-demand-disruption-60}
\end{figure}

The results are shown in Table~\ref{tab:customer-demand-disruption}. We see that for both MLP- and NAM-PPO the IQM of the reward degrades very similarly as we increase the disruption strength. In Figure~\ref{fig:customer-demand-disruption-60} we observe the same behaviour as reflected in the per-step reward, where the change in demand affects both MLP- and NAM-based agents.
    Overall, we find that both MLP- and NAM-PPO agents are similarly affected by changes in the demand distribution. 

Finally, we retrained the same PPO agents but this time on environments in which we introduce the same kind of disturbances ($s_{\textrm{d}}$ = 1, $c$ = 30) This can be interpreted as a form of hardening the RL policies towards these disruptions. Our preliminary results were collected over 5 random seeds with 50 evaluations each and are shown in Table~\ref{tab:customer-demand-disruption} (denoted as \emph{hardened}). They indicate that the disruptions in the environment made both the MLP- and NAM-based PPO policies more brittle than without introducing them.
That is, the hardening decreases the performance and makes the policies more susceptible towards the changes in the demand. 
Yet, MLP-PPO performs noticeably better than the NAM-based policy.
Overall, these results are surprising and require a more detailed investigation in future works.
    
\begin{table}
\centering
\small
\begin{tabular}{@{}S[table-format=1.1]llll@{}}
        \toprule
        {Disruption} & 
        NAM &
        MLP & 
        NAM hardened & 
        MLP hardened \\
        {strength $s_{\textrm{d}}$} & 
         &
         & 
        (5 seeds) & 
        (5 seeds) \\ \midrule
        0.0                                                           & 792.69  {\tiny {[}779.96, 804.9{]}}  & \phantom{--}730.27 {\tiny {[}709.12, 750.38{]}}  & \phantom{--}320.81 {\tiny {[}259.13, 379.49{]}}    & \phantom{--}605.72 {\tiny {[}561.15, 644.60{]}}     \\
        0.5                                                              & 804.21  {\tiny {[}788.19, 819.74{]}} & \phantom{--}718.28 {\tiny {[}690.88, 743.25{]}}   & \phantom{--}237.04 {\tiny {[}168.71, 303.13{]}}    & \phantom{--}556.71 {\tiny {[}501.31, 606.07{]}}    \\
        1.0                                                              & 727.32  {\tiny {[}707.34, 746.91{]}} & \phantom{--}629.68 {\tiny {[}599.39, 658.15{]}}   & \phantom{--}117.03 {\tiny {[}46.74, 185.30{]}}      & \phantom{--}451.96 {\tiny {[}391.87, 506.09{]}}    \\
        2.0                                                              & 490.14  {\tiny {[}469.41, 510.89{]}} & \phantom{--}388.06 {\tiny {[}358.38, 416.84{]}}   & --139.57 {\tiny {[}--209.49, --72.15{]}}  & \phantom{--}203.02 {\tiny {[}141.43, 260.71{]}}    \\
        4.0                                                              & --26.96 {\tiny {[}--48.42, --5.61{]}}   & --126.09 {\tiny {[}--155.18, --98.26{]}} & --661.41 {\tiny {[}--731.12, --593.59{]}} & --320.04 {\tiny {[}--382.29, -261.65{]}} \\ \bottomrule
\end{tabular}
\caption{Evaluation of the robustness towards disruptions in the demand distribution for 60 environment steps. We increase the disruption strength $s_d$ and show the IQM of the total reward. The experiment was carried out for each of the 20 models and over 50 rollouts each. The 5\% and 95\% bootstrap confidence intervals of the IQM are given in brackets.}
\label{tab:customer-demand-disruption}
\end{table}

\newpage
\subsection{Hyperparameter Details and Optimization}\label{app:sec:hyperparameter_optimization}

The hyperparameters and search space for hyperparameter optimization used throughout the provided experiments are summarized in Table~\ref{tab:hyperparameter_search_space}.

\begin{table}[ht]
    \centering
    \small
    \begin{tabular}{@{}llll@{}}
        \toprule
        Type & Hyperparameter                              & Value / Range            & Scaling   \\ \midrule
        Optimizer  & Learning rate                               & $[10^{-4}, 10^{-3}]$ & logarithmic  \\[0.75mm]
                   & Batch size                                & $[32, 128]$   & linear \\[2mm]
        Actor MLP       & \# of hidden layers                          & $[1, 4]$       & linear \\[0.75mm]
                    &    \# of neurons per layer              & $[8, 32]$      & linear \\[0.75mm]
                     &    Activation functions              & ELU      & fixed \\[2mm]
        Actor NAM       & \# of hidden layers                          & $[1, 4]$       & linear \\[0.75mm]
                    &    \# of neurons per layer              & $[8, 32]$      & linear \\[0.75mm]
                    &    \# of subnets $S$               & $30$      & fixed \\[0.75mm]
                    &    Hidden activation functions              & ELU      & fixed \\[2mm]
        Critic MLP       & \# of layers                          & $2$       & fixed \\[0.75mm]
                    &    \# of neurons per layer              & $64$      & fixed \\[0.75mm]
                     &    Activation functions              & ELU      & fixed \\[2mm]
        PPO     & \# of epochs                                & $[2, 51]$   & linear \\[0.75mm]
             & \# of steps                        & $2048$      & fixed \\[0.75mm]
             & Entropy coefficient                         & $0.01$      & fixed \\[0.75mm]
                & Gamma                         & $0.99$      & fixed \\\bottomrule
    \end{tabular}
    \caption{Hyperparameters used in all experiments. Hyperparameter optimization via Random Search is performed on non-fixed parameters in the specified value range. The search space is chosen comparably small in order to not introduce immense computational overhead.}
    \label{tab:hyperparameter_search_space}
\end{table}

\end{document}